\documentclass[10pt,onecolumn]{article}

\usepackage[utf8]{inputenc} 
\usepackage{longtable}
\usepackage{bm}
\usepackage{amsmath}
\usepackage{amssymb}
\usepackage{amsthm}
\usepackage{lmodern} 
\usepackage{fix-cm}  

\usepackage[dvips]{graphicx}
\usepackage{algorithm}
\usepackage{algorithmic}
\renewcommand{\algorithmicrequire}{\textbf{Input:}}
\renewcommand{\algorithmicensure}{\textbf{Output:}}
\theoremstyle{definition}

\theoremstyle{definition}

\theoremstyle{definition}

\theoremstyle{definition}

\theoremstyle{definition}

\theoremstyle{theorem}

\usepackage{color}
\definecolor{red}{rgb}{1,0,0}
\definecolor{blue}{rgb}{0,0,1}

\newcommand{\0}{{\bm{0}}}
\newcommand{\1}{{\bm{1}}}
\newcommand{\va}{{\bm{a}}}
\newcommand{\vb}{{\bm{b}}}
\newcommand{\vc}{{\bm{c}}}
\newcommand{\vd}{{\bm{d}}}

\newcommand{\x}{{\bm{x}}}
\newcommand{\y}{{\bm{y}}}

\newcommand{\vA}{{\bm{A}}}

\newcommand{\cC}{{\mathcal{C}}}
\newcommand{\vD}{{\bm{D}}}

\newcommand{\vF}{{\bm{F}}}

\newcommand{\vI}{{\bm{I}}}
\newcommand{\cI}{{\mathcal{I}}}

\newcommand{\bN}{{\mathbb{N}}}

\newcommand{\bO}{{\mathbb{O}}}

\newcommand{\vO}{{\bm{O}}}

\newcommand{\bR}{{\mathbb{R}}}

\newcommand{\bS}{{\mathbb{S}}}

\newcommand{\vU}{{\bm{U}}}

\newcommand{\vW}{{\bm{W}}}

\newcommand{\X}{{\bm{X}}}

\newcommand{\Y}{{\bm{Y}}}

\newcommand{\vxi}{{\bm{\xi}}}

\newcommand{\vPhi}{{\bm{\Phi}}}

\newcommand{\vThet}{{\bm{\Theta}}}

\DeclareMathOperator{\diag}{diag}

\DeclareMathOperator{\logm}{logm}

\newcommand{\argmin}{\mathop{\textrm{argmin}}\limits}

\begin{document}

\begin{center}
{\Large
  Stochastic Dykstra Algorithms for Metric Learning
  on Positive Semi-Definite Cone}
\\
\vspace{1cm}
       {\large
         Tomoki Matsuzawa${}^{\dagger}$, 
         Raissa Relator${}^{\diamond}$,
         Jun Sese${}^{\diamond}$, 
         Tsuyoshi Kato${}^{\dagger,\ddagger,*}$}
\\
\vspace{1cm}
\begin{tabular}{lp{0.7\textwidth}}
${}^\dagger$ & 
Faculty of Science and Engineering, Gunma University, 
Kiryu-shi, Gunma, 326--0338, Japan.  
\\
${}^\ddagger$ &
Center for Informational Biology, Ochanomizu University, 
Bunkyo-ku, Tokyo,
112--8610, Japan. n
\\
${}^\diamond$ &
BRD, AIST, Koto-ku, Tokyo, 135--0064, Japan.
\\
\end{tabular}
\end{center}

\nocite{KatTsuAsa05a,RelNagKat16a}

\begin{abstract}
Recently, covariance descriptors have received much attention as powerful representations of set of points. In this research, we present a new metric learning algorithm for covariance descriptors based on the Dykstra algorithm, in which the current solution is projected onto a half-space at each iteration, and runs at $O(n^3)$ time. We empirically demonstrate that randomizing the order of half-spaces in our Dykstra-based algorithm significantly accelerates the convergence to the optimal solution. Furthermore, we show that our approach yields promising experimental results on pattern recognition tasks.
\end{abstract}
\section{Introduction}
Learning with example objects characterized by a set of several points, instead of a single point, in a feature space is an important task in the computer vision and pattern recognition community. In the case of visual categorization of still images, many local image descriptors such as SIFT~\cite{lowe2004distinctive} are extracted from an input image to form a single vector such as a Bag-of-Visual-Words vector or a Fisher Vector~\cite{MatRelTak15a,perronnin2010improving}. For image set classification, a surge of methods have been developed in the last decade, and probabilistic models
\cite{shakhnarovich2002face} or kernels~\cite{RelHirItoKat14a} are introduced to describe the image set. Alternative descriptors are the covariance descriptors, which have received much attention as a powerful representation of a set of points.

The performance of categorizing covariance descriptors depends on the metric that is used to measure the distances between them.
To compare covariance descriptors, a variety of distance measures such as affine invariant Riemannian metric~\cite{pennec2006riemannian}, Stein metric~\cite{sra2012a}, J-divergence~\cite{wang2004affine}, Frobenius distance~\cite{JayasumanaCVPR13}, and Log-Frobenius distance~\cite{arsigny2006log}, have been discussed in existing literature. Some of them are designed from their geometrical properties, but some are not. Many of these distance measures are expressed in the form
\begin{align*}
D_{\vPhi}(\X_{1},\X_{2}) 
:=
\left\|\vPhi(\X_{1})-\vPhi(\X_{2})\right\|^{2}_{\text{F}},    
\end{align*}
with $\vPhi:\bS_{++}^{n}\to\bR^{n\times m}$ for some $m\in\bN$.
If $\vPhi(\X):=\logm(\X)$, where $\logm(\X)$ takes the principal matrix logarithm of a strictly positive definite matrix $\X$, the 
Log-Frobenius distance~\cite{arsigny2006log} is obtained. Setting $\vPhi(\X):=\X^{p}$ gives the Power-Frobenius distance~\cite{JayasumanaCVPR13}, while $\vPhi(\X):=\text{chol}(\X)$, where $\text{chol}:\bS^{n}_{+}\to\bR^{n\times n}$ produces the Cholesky decomposition of $\X$ such that $\X = \text{chol}(\X)\text{chol}(\X)^\top$, yields the Cholesky-Frobenius distance~\cite{Dryden09a}. These metrics are pre-defined before the employment of machine learning algorithms, and are not adaptive to the data to be analyzed. Meanwhile, for categorization of vectorial data, supervised learning for fitting metrics to the task has been proven to significantly increase the performance of the distance-based classifier~\cite{DavKulJaiSraDhi07a, KatNag10a,RelNagKat16a}.

In this paper, we introduce a parametric distance measure between covariance descriptors and present novel metric learning algorithms to determine the parameters of the distance measure function. The learning problem is formulated as the Bregman projection onto the intersections of half-spaces. This kind of problem can be solved by the Dykstra algorithm~\cite{Censor98,Dykstra83}, which chooses a single half-space in a cyclic order and projects a current solution to the half-space. We developed an efficient technique for projection onto a single half-space. Furthermore, we empirically found that selecting the half-space stochastically, rather than in a cyclic order, dramatically increases the speed of converging to an optimal solution.

\subsection{Related work}
To the best of our knowledge, Vemulapalli et al. (2015)~\cite{vemulapalli2015riemannian}
were the first to introduce the supervised metric
learning approach for covariance descriptors. They vectorized the matrix logarithms of the covariance descriptors to apply existing metric learning methods to the vectorizations of
matrices. The dimensionality of the vectorizations is
$n(n+1)/2$ when the size of the covariance matrices are
$n\times n$. Thus, the size of the Mahalanobis matrix is
$n(n+1)/2\times n(n+1)/2$, which is computationally
prohibitive when $n$ is large.

Our approach is an extension of the distance measure of Huang et al.~\cite{ZhiwuHuang15a}, which is based on the Log-Euclidean metric, with their loss function being a special case of our formulation. They also adopted the cyclic Dykstra algorithm for learning the Mahalanobis-like matrix. However, they misused the Woodbury matrix inversion formula when deriving the projection onto a single half-space, therefore, their algorithm has no theoretical guarantee of converging to the optimal solution. In this paper, their update rule is corrected by presenting a new technique that projects a current solution to a single half-space within $O(n^{3})$ computational time.

Yger and Sugiyama~\cite{yger2015supervised} devised a different formulation of metric learning. They introduced the congruent transform~\cite{Bhatia-book09} and measures distances between the transformations of covariance
descriptors. An objective function based on the kernel target alignment~\cite{Cristianini01kta} is employed to determine the transformation parameters. Compared to their algorithm, our algorithm has the capability to monitor the upper bound of the objective gap, i.e. the difference between the current objective and the minimum. This implies that the resultant solution is ensured to be $\epsilon$-suboptimal if the algorithm’s convergence criterion is set such that the objective gap upper bound is less than a very small number $\epsilon$. Since Yger and Sugiyama~\cite{yger2015supervised}  employed a gradient method for learning the congruent transform, there is no way to know the objective gap.

\subsection{Contributions}
Our contributions of this paper can be
summarized as follows.
\begin{itemize}
  \item For metric learning on positive semidefinite cone, we developed a new algorithm based on the Dykstra algorithm, in which the current solution is projected onto a half-space at each iterate, and runs at $O(n^{3})$ time.
\item
We present an upper-bound for the objective gap which provides a stopping criterion and ensures the optimality of the solution.
\item
  We empirically found that randomizing the order of half-spaces in our Dykstra-based algorithm significantly accelerates the convergence to the optimal solution.
\item
We show that our approach yields promising experimental results on pattern recognition tasks.
\end{itemize}

\subsection{Notation}
We denote vectors by bold-faced lower-case letters and
matrices by bold-faced upper-case letters.
Entries of vectors and matrices are not bold-faced.
The transposition of a matrix $\vA$ is denoted by $\vA^{\top}$,
and the inverse of $\vA$ is by $\vA^{-1}$.
The $n\times n$ identity matrix is denoted by $\vI_{n}$. 
The subscript is often omitted. 
The $m\times n$ zero matrix is denoted by $\vO_{m\times n}$. 
The subscript is often omitted. 
%
The $n$-dimensional vector all of whose entries are one is denoted by $\1_{n}$.
We use $\bR$ and $\bN$ to denote the set of real and natural numbers,
$\bR^{n}$ and $\bN^{n}$ to denote the set of $n$-dimensional real and natural vectors,
and $\bR^{m\times n}$ to denote the set of $m\times n$ real matrices.
For any $n\in\bN$, we use $\bN_{n}$ to denote the set of natural numbers less than or equal to $n$.
%
%
%
%
Let us define 
$\bR_{+}:=\{ x\in\bR \,|\, x\ge 0 \}$,
$\bR_{++}:=\{ x\in\bR \,|\, x> 0 \}$,  
$\bR_{+}^{n}:=\{ x\in\bR^{n} \,|\, \x\ge \0_{p} \}$, and 
$\bR_{++}^{n}:=\{ x\in\bR^{n} \,|\, \x> \0_{p} \}$. 
The relational operator $\succ$ denotes the generalized inequality
associated with the strictly positive definite cone.
%
We use
$\bS^{n}$ to denote the set of symmetric $n\times n$ matrices.
$\bS_{+}^{n}$ to denote the set of symmetric positive semi-definite $n\times n$ matrices,
and
$\bS_{++}^{n}$ to denote the set of symmetric strictly positive definite $n\times n$ matrices.
%
%
%
For any $\x = \left[x_{1},\dots,x_{n}\right]^\top\in\bR^{n}$, 
$\text{diag}(\x)$ is defined as an $n\times n$ 
diagonal matrix whose diagonal entries are 
$x_{1},\dots,x_{n}$. 
For any $n\times n$ square matrix $\X$, its trace is denoted by 
$\text{tr}(\X)$. 
For any $\x,\y\in\bR^{n}$, define $\left<\x,\y\right> := \sum_{i=1}^{n}x_{i}y_{i}$
where $x_{i}$ and $y_{i}$ is the $i$-th entry of $\x$ and $\y$, respectively. 
For any $\X,\Y\in\bR^{m\times n}$, 
define $\left<\X,\Y\right> := \sum_{i=1}^{m}\sum_{j=1}^{n}X_{i,j}Y_{i,j}$
where $X_{i,j}$ and $Y_{i,j}$ is the $(i,j)$-th entry of $\X$ and $\Y$, respectively.
%
%
$\bO_{n}$ is used to denote the set of $n\times n$ orthonormal
matrices, i.e.
$\bO_{n} := \{\vA \in\bR^{n\times n} \,|\, \vA^\top \vA = \vI_{n}\}.$ 
%
\section{Our Metric Learning Problem}
\subsection{Parametric distance measure on $\bS_{+}^{n}$}
We introduce the following distance measure for covariance descriptors $\X_{1},\X_{2}\in\bS_{+}^{n}$:
\begin{align*}
D_{\vPhi}(\X_{1},\X_{2};\vW) 
:=
\left<\vW,\left(\vPhi(\X_{1})-\vPhi(\X_{2})\right)
\left(\vPhi(\X_{1})-\vPhi(\X_{2})\right)^\top
\right>,
\end{align*}
where $\vW\in\bS_{+}^{n}$ is the parameter of this distance measure function. If $\vW$ is strictly positive definite and  $\vPhi$ is bijective, then this distance measure $D_{\vPhi}(\cdot,\cdot;\vW):\bS_{+}^{n}\times \bS_{+}^{n}\to\bR$ is a metric because all of the following conditions are satisfied:%
(i) non-negativity:
$D_{\vPhi}(\X_{1},\X_{2};\vW) \ge 0; $
(ii) identity of indiscernibles:
$D_{\vPhi}(\X_{1},\X_{2};\vW) = 0 \text{ iff } \X_{1}=\X_{2};$
(iii) symmetry:
$D_{\vPhi}(\X_{1},\X_{2};\vW) = D_{\vPhi}(\X_{2},\X_{1};\vW);$
(iv) triangle inequality:
$D_{\vPhi}(\X_{1},\X_{3};\vW) \le
D_{\vPhi}(\X_{1},\X_{2};\vW) + D_{\vPhi}(\X_{2},\X_{3};\vW).$
If the parameter matrix $\vW$ is singular, $D_{\vPhi}(\cdot,\cdot;\vW)$ is a pseudometric, and the identity of indiscernibles is changed to the following property:  
For any $\X_{1}\in\bS_{++}^{n}$, $D_{\vPhi}(\X_{1},\X_{1};\vW) = 0$ holds, while $D_{\vPhi}(\X_{1},\X_{2};\vW) = 0$ occurs for some non-identical positive semi-definite matrices $\X_{1}$ and $\X_{2}$.  

\subsection{Formulations of the learning problems}
To determine the value of the parameter matrix~$\vW$, we pose a constrained optimization problem based on the idea of ITML~\cite{DavKulJaiSraDhi07a}. We now consider a multi-class categorization problem. Let $n_{c}$ be the number of classes, and the class labels are represented by natural numbers in $\bN_{n_{c}}$. Suppose weare given%
\begin{align*}
  (\X_{1},\omega_{1}),\dots,(\X_{\ell},\omega_{\ell})\in\bS_{+}^{n}\times\bN_{n_{c}}
\end{align*}
as a training dataset, where $\X_{i}$ is
the covariance descriptor of the $i$-th example, and
$\omega_{i}$ is its class label. From the $\ell$ examples,
$K$ pairs
$(i_{1},j_{1}),\dots,(i_{K},j_{K})\in\bN_{\ell}\times\bN_{\ell}$
are picked to give, to each pair, the
following constraint:
\begin{align}\label{eq:con01}
D_{\vPhi}(\X_{i_{k}},\X_{j_{k}};\vW) 
\begin{cases}
\le b_{\text{ub}}\xi_{k},
&\qquad \text{if $\omega_{i_{k}}=\omega_{j_{k}}$}, 
\\
\ge b_{\text{lb}}\xi_{k},
&\qquad \text{if $\omega_{i_{k}}\ne \omega_{j_{k}}$},  
\end{cases}
\end{align}
where, when $\xi_{k}=1$,
the two constants~$b_{\text{ub}}$ and $b_{\text{lb}}$,
respectively, are the upper-bound of the distances between any two examples in the same class and the lower-bound of the distances between any two examples in different classes.
Now let us define for $k\in\bN_{K}$, 
\begin{align*}
  y_{k} :=
  \begin{cases}
    +1, \qquad \text{if }\omega_{i_{k}}=\omega_{j_{k}},
    \\
    -1, \qquad \text{if }\omega_{i_{k}}\ne\omega_{j_{k}},
  \end{cases}
  \qquad
  \text{and}
  \qquad
  b_{k} :=
  \begin{cases}
    b_{\text{ub}}, \qquad \text{if }\omega_{i_{k}}=\omega_{j_{k}},
    \\
    b_{\text{lb}}, \qquad \text{if }\omega_{i_{k}}\ne\omega_{j_{k}}.
  \end{cases}
\end{align*}
Under the constraint \eqref{eq:con01}, we wish to find $\vW$ and $\xi_{k}$ such that $\vW$ is not much deviated from the identity matrix and $\xi_{k}$ is close to one. From this motivation, we pose the following problem:
\begin{align}\label{eq:soft-covitml-breg}
\text{min }\quad&
\text{BD}_{\varphi}((\vW,\vxi),(\vI,\1)) , 
\qquad
\text{wrt }\quad \vW\in\bS_{++}^{n}, \quad
\vxi = \left[\xi_{1},\dots,\xi_{K}\right]^\top \in\bR_{++}^{K}, 
\\
\text{subject to }\quad&
\forall k\in\bN_{K}, \quad
y_{k} D_{\vPhi}(\X_{i_{k}},\X_{j_{k}};\vW) 
\le y_{k} b_{k}\xi_{k}, 
\nonumber 
\end{align}
where
$\text{BD}_{\varphi}(\cdot,\cdot):(\bS_{++}^{n}\times\bR_{++}^{K})\times (\bS_{++}^{n}\times\bR_{++}^{K})\to\bR_{+}$
is the \emph{Bregman divergence}~\cite{KatTakOma13a}. 
Only if $(\vW,\vxi)=(\vI,\1)$ will the divergence
$\text{BD}_{\varphi}((\vW,\vxi),(\vI,\1))$ become zero,
and the value of divergence becomes larger if
$(\vW,\vxi)$ is more deviated from $(\vI,\1)$.
The definition of the Bregman divergence contains a seed function $\varphi: \bS_{++}^{n}\times \bR_{++}^{K}\to \bR$ which is assumed to be continuously differentiable and strictly convex.  For some $\varphi$, the Bregman divergence is defined as
\begin{align*}
\text{BD}_{\varphi}(\vThet,\vThet_{0})
= \varphi(\vThet) - \varphi(\vThet_{0}) - 
\left< \nabla \varphi(\vThet_{0}), \vThet-\vThet_{0} \right>, 
\end{align*}
for
$\vThet,\vThet_{0}\in\bS_{++}^{n}\times\bR_{++}^{K}$.
This implies that the quantities of the deviations of
the solution $(\vW,\vxi)$ from $(\vI,\1)$ depend on
the definition of the seed function. In this study,
the seed function is assumed to be the sum of two terms:
\begin{align*}
  \varphi( \vW, \vxi )
  := \varphi_{\text{r}}(\vW) +
  \sum_{k=1}^{K}c_{k}\varphi_{\text{l}}(\xi_{k}), 
\end{align*}
where $c_{k}$ is a positive constant. 
The first term $\varphi_{\text{r}}:\bS_{++}^{n}\to\bR$ in the definition of the seed function is defined by $\varphi_{\text{r}}(\vW):=- \text{logdet}(\vW)$.As for the definition of the second term$\varphi_{\text{l}}:\bR_{++}\to\bR$, we considered the following three functions:
\begin{align*}
\varphi_{\text{is}}(\xi_{k}) &:= -\log(\xi_{k}), &
\varphi_{\text{l2}}(\xi_{k}) &:= \frac{1}{2}\xi_{k}^{2}, &
\varphi_{\text{e}}(\xi_{k}) &:= (\log \xi_{k} - 1)\xi_{k}.
\end{align*}
The Bregman divergences generated from three seed functions~$\varphi_{\text{is}}$, $\varphi_{\text{l2}}$, and $\varphi_{\text{e}}$, respectively, are referred to as \emph{Itakura-Saito Bregman Divergence} (ISBD), \emph{L2 Bregman Divergence} (L2BD), and \emph{Relative Entropy Bregman Divergence} (REBD), where ISBD is equal to the objective function employed by Huang et al.~\cite{ZhiwuHuang15a}.
\section{Stochastic Variants of Dykstra Algorithm}
We introduce the Dykstra algorithm~\cite{Censor98,Dykstra83}
to solve the optimization problem~\eqref{eq:soft-covitml-breg}.
The original Dykstra algorithm~\cite{Dykstra83}
was developed as a computational method that finds the
Euclidean projection from a point onto the intersection of
convex sets. Censor \& Reich~\cite{Censor98} extended the algorithm
to finding the Bregman projection from a point $\x_{0}$ to a set $\cC$,
defined by
\begin{align*}
\argmin_{\x\in\cC}\text{BD}_{\varphi}(\x,\x_{0}). 
\end{align*}

In available literature related to stochastic gradient
descent methods and the
variants~\cite{Bottou10a-sgd,Johnson13a-svrg,Roux12a-sag,Shalev-Shwartz11a-pegasos}
that minimize the regularized loss averaged
over a set of examples, it is empirically
shown that, rather than picking an example in a cyclic order, example selection in a stochastic order dramatically speeds up the
convergence to the optimal solution. Alternatively, some
literature reported that at the beginning of every epoch in
the gradient method, random permutation of the order of
examples also accelerates the convergence~\cite{defazio2014finito}.

Motivated by these facts, this study proposes the use of
stochastic orders for selection of convex set components in
the Dykstra algorithm. We term the stochastic version of the
Dykstra algorithm as the \emph{stochastic Dykstra algorithm}. In
our case, every convex set component is one of $K$
half-spaces, as will be described in a later discussion. There are, then, three ways to
select half-spaces:
\begin{itemize}
\item \textbf{Cyclic}:
  Pick a half-space in a cyclic order at each iteration.
\item \textbf{Rand}:
  Pick a half-space randomly at each iteration.
\item \textbf{Perm}:
  Permute the order of $K$ half-spaces randomly at the
  beginning of each epoch.
\end{itemize}
Hereinafter, we assume to employ the ``Rand'' option,
although replacing this option with one of the remaining two is
straightforward.

If every convex set component is a half-space, and the
$k$-th convex set component~$\cC_{k}$ is expressed as
\begin{align*}
  \cC_{k} := \left\{ \x \,|\, \left<\va_{k},\x\right> \le b_{k} \right\}, 
\end{align*}
then computing the Bregman projection from a point $\x_{0}$
to its boundary $\text{bd}(\cC_{k})$ is equivalent to
solving the following saddle point problem:
\begin{align*}
\max_{\delta} \min_{\x}
\text{BD}_{\varphi}(\x,\x_{0}) + \delta ( \left<\va_{k},\x\right> - b_{k} ). 
\end{align*}
This fact enables us to rewrite the Dykstra algorithm
with Rand option for finding the Bregman projection
from a point $\x_{0}$ to the intersection of
$\cC_{1},\dots,\cC_{K}$, as described in Algorithm~\ref{algo:dykstra},
where $\varphi^{*}$ is the convex conjugate of the seed function $\varphi$. 
\begin{algorithm}[th!]
\caption{
Stochastic Dykstra Algorithm. 
\label{algo:dykstra}}
\begin{algorithmic}[1]
\STATE \textbf{begin}
\STATE $\forall k\in\bN_{K}:\; \alpha_{k} := 0;$
\FOR{$t = 1, 2, \dots$}
\STATE Pick $k$ randomly from $\{1,\dots,K\}$;  
\STATE Solve the following saddle point problem
and let $\delta_{t-1/2}$ be the solution of $\delta$: 
\begin{align}\label{eq:saddleprob-general}
\max_{\delta} \min_{\x}
\text{BD}_{\varphi}(\x,\x_{t-1}) + \delta_{t} ( \left<\va_{k},\x\right> - b_{k} );   
\end{align}
\STATE 
$\delta_{t} := \max( \delta_{t-1/2}, -\alpha_{k} ); $
$\alpha_{k} := \alpha_{k} + \delta_{t}; $
\STATE 
$\x_{t} = \nabla\varphi^{*}(\nabla\varphi(\x_{t-1}) - \delta_{t}\va_{k})$; 
\ENDFOR
\STATE \textbf{end.}
\end{algorithmic}
\end{algorithm}

\section{Efficient Projection Technique}
We now show that
solving the optimization problem~\eqref{eq:soft-covitml-breg}
is equivalent to finding a Bregman projection from a point
$(\vI,\1)\in\bS_{++}^{n}\times\bR_{++}^{K}$
onto the intersection of multiple half-spaces.

Let $\vA_{k}$ be a positive semidefinite matrix expressed as 
\begin{align*}
\vA_{k} := 
\left(\vPhi(\X_{i_{k}})-\vPhi(\X_{j_{k}})\right)
\left(\vPhi(\X_{i_{k}})-\vPhi(\X_{j_{k}})\right)^{\top}
\end{align*}
for $k\in\bN_{K}$, to define a half-space
\begin{align*}
\cC_{k} :=
\left\{
(\vW,\vxi)\in\bS_{++}^{n}\times \bR_{++}^{K}\,|\,
y_{k}\left<\vA_{k},\vW\right> - y_{k}b_{k}\xi_{k} \le 0
\right\}. 
\end{align*}
Then, it can be seen that the intersection of $K$ half-spaces
\begin{align*}
  \bigcap_{k=1}^{K}\cC_{k}
\end{align*}
is the feasible region of 
the optimization problem~\eqref{eq:soft-covitml-breg}.
This implies that the Dykstra algorithm can be applied to
solve problem~\eqref{eq:soft-covitml-breg}. 

Next we present an efficient technique that projects
$(\vW_{t-1},\vxi_{t-1})\in\bS_{++}^{n}\times\bR_{++}^{K}$
onto the $k$-th half-space $\cC_{k}$, 
where $(\vW_{t-1},\vxi_{t-1})\in\bS_{++}^{n}\times\bR_{++}^{K}$ is 
the model parameter after the $(t-1)$-th iteration.  
Let $\xi_{k,t-1}$ be the $k$-th entry in the vector~$\vxi_{t-1}$.
The value of the function $J_{t}:\bR\to\bR$ defined by
\begin{align*}
  J_{t}(\delta) :=
  \left< \vA_{k}, (\vW_{t-1}^{-1}+\delta y_{k}\vA_{k})^{-1}\right>
  - b_{k}
  \nabla\varphi^{*}_{\text{l}}
  (\nabla\varphi_{\text{l}}(\xi_{t-1}) + \delta y_{k}b_{k}/c_{k}), 
\end{align*}
is zero at the solution $\delta_{t-1/2}$
of the saddle point problem~\eqref{eq:saddleprob-general}.
The solution $\delta$ must satisfy
the strictly positive definiteness:
\begin{align}\label{eq:W-gt-zero}
  (\Y(\delta))^{-1} := \vW_{t-1}^{-1}+\delta y_{k}\vA_{k} \succ \vO,
\end{align}
and the feasibility of the slack variables: 
\begin{align}\label{eq:xi-in-ridom}
  \exists \, \xi_{k,t-1/2} \quad\text{s.t.}\quad
  \nabla\varphi_{\text{l}}(\xi_{k,t-1/2}) =
  \nabla\varphi_{\text{l}}(\xi_{k,t-1}) - \delta y_{k}b_{k}/c_{k} . 
\end{align}
There is no closed-form solution found for this projection
problem.  Hence, some numerical method such as the Newton-Raphson
method is necessary for solving the nonlinear system $J_{t}(\delta)=0$.
If one tries to compute the value of $J_{t}(\delta)$ na\"{i}vely,
it will require an $O(n^{3})$ computational cost because 
$J_{t}(\cdot)$ involves computation of the inverse of
an $n\times n$ matrix. 
If we suppose the numerical method assesses the value of
the scalar-valued function $J_{t}(\cdot)$ $L$ times, the
na\"{i}ve approach will take $O(Ln^{3})$ computational time
to find the solution of the nonlinear system $J_{t}(\delta)=0$.
Furthermore, the positive definiteness condition in~\eqref{eq:W-gt-zero}
and the feasibility condition in~\eqref{eq:xi-in-ridom} must be checked.

We will show the following two claims:
\begin{itemize}
\item The solution of the
system $J_{t}(\delta)=0$ satisfying \eqref{eq:W-gt-zero}
and \eqref{eq:xi-in-ridom} can be
computed within $O(n^{3}+Ln)$ time, where $L$ is
the number of times a numerical method assesses the value of
$J_{t}(\cdot)$.
\\
\item The solution exists and it is unique.
\end{itemize}
Hereinafter, we assume $\vA_{k}$ is strictly positive definite.
By setting
$\vA_{k} \leftarrow \vA_{k} + \epsilon \vI$, with $\epsilon$ as a small positive
constant, it is easy to satisfy this assumption.
Since $L \in O(n^{2})$ in a typical setting, we can say that
each update can be done in $O(n^{3})$ computation. 

We define $\vA^{1/2}_{k}$, $\vU$, $\vD$, and $\vd$ as follows. 
Let $\vA_{k}^{1/2}\in\bS_{++}^{n}$ 
such that $\vA_{k}^{1/2}\vA_{k}^{1/2}=\vA_{k}$, and 
denote by $\vA_{k}^{-1/2}\in\bS_{++}^{n}$ the inverse of $\vA_{k}^{1/2}$.
Introduce an orthonormal matrix $\vU\in\bO_{n}$ and 
a diagonal matrix $\vD=\diag(\{d_{1},\dots,d_{n}\})$ that 
represent a spectral decomposition 
$\vU\vD\vU^{\top} = \vA^{-1/2}\vW_{t-1}^{-1}\vA^{-1/2}$,
with $d_{1}\ge\dots\ge d_{n}$. 
Then, we have
\begin{align}\label{eq:Y-diagnolized}
\Y(\delta) = 
\vA_{k}^{-1/2} \vU
(\vD + y_{k} \delta \vI)^{-1} \vU^{\top} \vA_{k}^{-1/2},
\end{align}
which allows us to rewrite the first term of $J_{t}(\delta)$ as 
\begin{align}\label{eq:dotprod-linear-comp}
\left<\vA_k,
(\vW_{t-1}^{-1} + \delta y_{k} \vA_k)^{-1}
\right> 
=
\sum_{i=1}^{n}\frac{1}{d_{i}+y_{k}\delta}. 
\end{align}
Assessment of $J_{t}(\delta)$ can be
done within $O(n)$ computational cost after $d_{1},\dots,d_{n}$ are obtained.
To get the $n$ scalars $d_{1},\dots,d_{n}$,
we need to find $\vA_k^{-1/2}$ and the
spectral decomposition of $\vA^{-1/2}_k\vW_{t-1}^{-1}\vA^{-1/2}_k$, 
each of which requires $O(n^{3})$ computation. 
The $n\times n$ matrix $\vA^{-1/2}_{k}$ can be computed
in the pre-process of the Dykstra algorithm, while 
the spectral decomposition of $\vA^{-1/2}_k\vW_{t-1}^{-1}\vA^{-1/2}_k$
is done once before invoking some numerical method to
solve the nonlinear system $J_{t}(\delta)=0$. These support the first claim.

Equation~\eqref{eq:Y-diagnolized} suggests that 
the set of $\delta$ satisfying \eqref{eq:W-gt-zero}
is given by
\begin{align}\label{eq:int-nlrsys-for-upd-alph}
I_{\text{r},t} := 
\begin{cases}
( -d_{n},+\infty ), &\qquad \text{for $y_{k}=+1$}, 
\\
( -\infty, d_{n} ), &\qquad \text{for $y_{k}=-1$}. 
\end{cases}
\end{align}

The set of $\delta$ satisfying \eqref{eq:xi-in-ridom} is given
as follows.  In the case of using ISBD,
$\delta$ ensuring \eqref{eq:xi-in-ridom} is in
the interval
\begin{align*}
I_{\text{is},t} := 
\begin{cases}
( -\infty, \delta_{\text{b}} ), &\qquad \text{for $y_{k}=+1$}, 
\\
( -\delta_{\text{b}}, +\infty ), &\qquad \text{for $y_{k}=-1$}, 
\end{cases}
\end{align*}
where 
\begin{align*}
  \delta_{\text{b}} := \frac{c_{k}}{b_{k}\xi_{k,t-1}}. 
\end{align*}
In the case of using L2BD and REBD, there exists $\xi_{t-1/2}$ even if
$\nabla\varphi_{\text{l}}(\xi_{t-1}) - \delta y_{k}b_{k}/c_{k}$ takes any value.

Hence, if ISBD is employed, the solution $\delta_{t-1/2}$ can be searched
from the interval
\begin{align*}
  I_{\text{r},t}\cap I_{\text{is},t} =
\begin{cases}
( -d_{n}, \delta_{\text{b}} ), &\qquad \text{for $y_{k}=+1$}, 
\\
( -\delta_{\text{b}}, +d_{n} ), &\qquad \text{for $y_{k}=-1$}. 
\end{cases}
\end{align*}
If L2BD or REBD is employed, 
the solution $\delta_{t-1/2}$ can be searched from $I_{\text{r},t}$.
In the reminder of this section, we shall use the notation $I_{t}$ to
denote the interval for $\delta$ satisfying \eqref{eq:W-gt-zero}
and \eqref{eq:xi-in-ridom} simultaneously.

We now show the uniqueness of the solution. The gradient of $J_{t}:\bR\to\bR$
is expressed as
\begin{align*}
  \nabla J_{t}(\delta)
  =
  -\sum_{i=1}^{n}\frac{y_{k}}{(d_{i}+y_{k}\delta)^{2}}
  -\frac{y_{k}b_{k}^{2}}{c_{k}}
  \nabla^{2}\varphi^{*}_{\text{l}}
  (\nabla\varphi_{\text{l}}(\xi_{t-1}) - \delta y_{k}b_{k}/c_{k}), 
\end{align*}
for $\delta\in I_{t}$.  We first consider the case that 
$y_{k}=+1$. Clearly, the first term is negative. 
The second term is non-positive because any convex conjugate function
is convex. Therefore, we have $\nabla J_{t}(\delta) < 0$.
In the case of $y_{k}=-1$, 
we get $\nabla J_{t}(\delta) > 0$ from a similar derivation. 
These observations imply that the solution is unique
if a solution exists.  The existence of the solution can be
established by showing that the curve $J_{t}(\delta)$ crosses
the horizontal axis.

We consider the cases of using ISBD and using either 
L2BD or REBD separately.
For the ISBD case, we have
\begin{align*}
 \lim_{\delta\searrow -d_{n}}J_{t}(\delta) &= +\infty,
 &
 \lim_{\delta\nearrow \delta_{\text{b}}}J_{t}(\delta) &= -\infty,
\end{align*}
if $y_{k}=+1$, and
\begin{align*}
 \lim_{\delta\searrow -\delta_{\text{b}}}J_{t}(\delta) &= -\infty,
 &
 \lim_{\delta\nearrow d_{n}}J_{t}(\delta) &= +\infty, 
\end{align*}
if $y_{k}=-1$.
On the other hand, when using either L2BD or REBD with $y_{k}=+1$ we get
\begin{align*}
 \lim_{\delta\searrow -d_{n}}J_{t}(\delta) &= +\infty,
 &
 \lim_{\delta\to +\infty}J_{t}(\delta) &= -\infty,
\end{align*}
while we obtain 
\begin{align*}
 \lim_{\delta\to -\infty}J_{t}(\delta) &= -\infty,
 &
 \lim_{\delta\nearrow d_{n}}J_{t}(\delta) &= +\infty, 
\end{align*}
when $y_{k}=-1$. Hence, we conclude that
\begin{align*}
 \exists ! \,\delta\in I_{t} \qquad \text{s.t.} \quad
 J_{t}(\delta) = 0. 
\end{align*}

\subsection{Stopping Criterion}
%
Here we discuss how to determine if the solution is already optimal and when to terminate the algorithm. 
While running the algorithm,
$(\vW_{t},\vxi_{t})$ may be infeasible to the primal
problem. Denote the index set of the violated constraints by
$\cI_{\text{vio}} := \{ k\in\bN_{K}\,|\, (\vW_{t},\vxi_{t})\not\in\cC_{k} \}$ 
and let us define $\bar{\vxi}_{t}\in\bR_{++}^{K}$ so that
the $k$-th entry is given by
$\bar{\xi}_{h,t} := \frac{1}{b_{h}}\left<\vW_{t},\vA_{h}\right>$
for $h\in\cI_{\text{vio}}$ and
$\bar{\xi}_{h,t} := \xi_{h,t}$ for $h\not\in\cI_{\text{vio}}$.
Note that $(\vW_{t},\bar{\vxi}_{t})$ is a feasible solution,
and $\bar{\vxi}_{t}=\vxi_{t}$ when $(\vW_{t},\vxi_{t})$ is feasible. 
The objective gap after iteration $t$
is bounded as follows:
\begin{align*}
  \text{BD}_{\varphi}((\vW_{t},\bar{\vxi}_{t}),(\vI,\1)) - \text{BD}_{\star}
  \le
  \sum_{h\in\cI_{\text{vio}}}
  c_{h}
  \left(
  \varphi_{\text{l}}(\bar{\xi}_{h,t})-\varphi_{\text{l}}({\xi}_{h,t})
  -\nabla\varphi_{\text{l}}(1)(\bar{\xi}_{h,t}-{\xi}_{h,t})
  \right)
  -
  \sum_{h=1}^{K}\alpha_{h}y_{h}
  \left(\left<\vA_{h},\vW_{t}\right> - b_{h}\xi_{h,t}\right), 
\end{align*}
where we have defined 
\begin{align*}
  \text{BD}_{\star} := \min_{(\vW,\vxi)\in\bigcap_{h}\cC_{h}}
  \text{BD}_{\varphi}((\vW,\vxi),(\vI,\1)). 
\end{align*}
Then this upper-bound of the objective gap can be used for the stopping
criterion of the Dykstra algorithm.   
\section{Numerical Experiments}

We conducted experiments to assess the convergence speed of
our optimization algorithms and the generalization
performance for pattern recognition.

\begin{figure*}[t!]
\begin{center}
\begin{tabular}{l}
\includegraphics[width=1.0\textwidth]{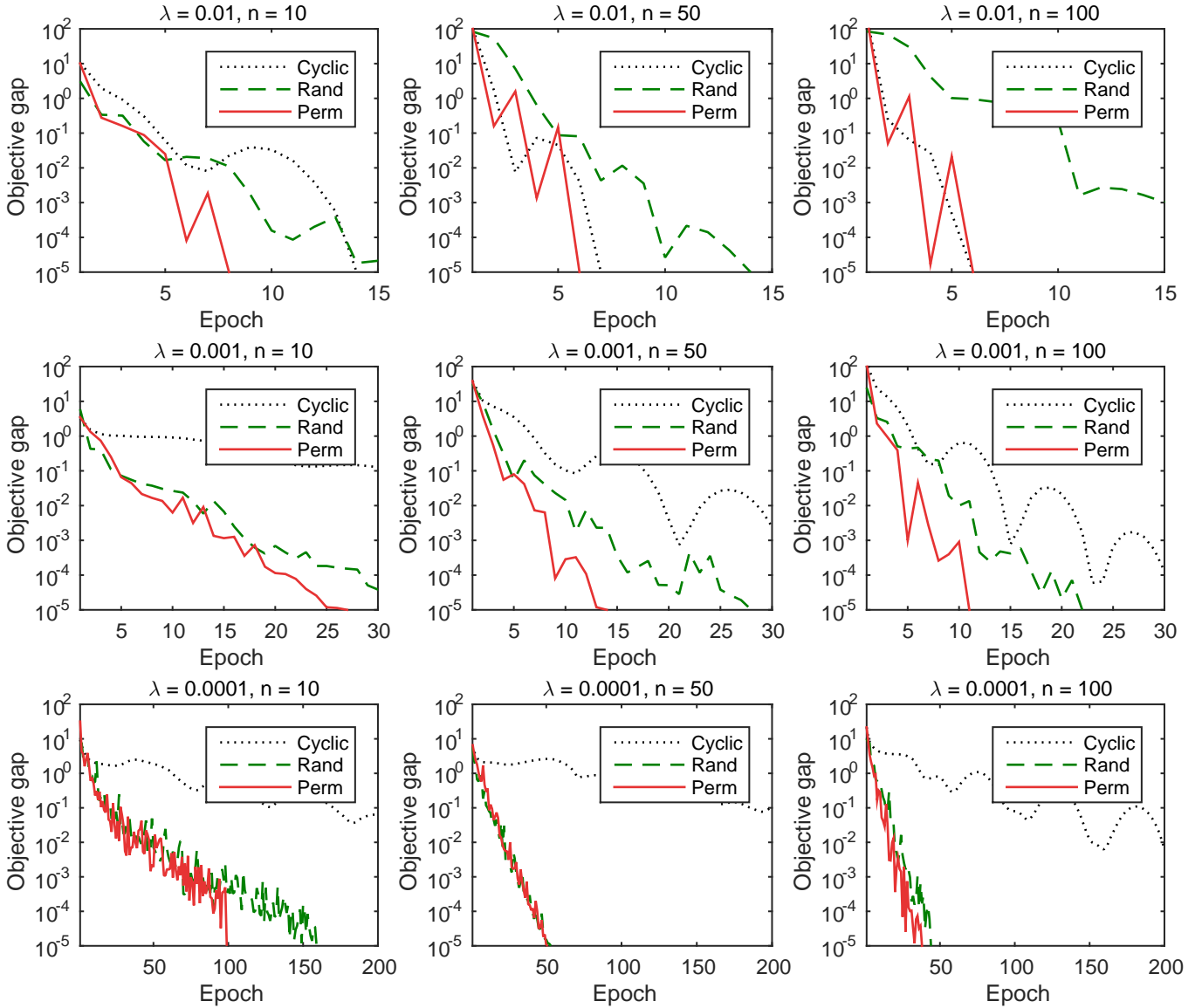}
\end{tabular}
\end{center}
\caption{Convergence behavior of the algorithms using different settings. \label{fig:logdet.primal}}
\end{figure*}

\subsection{Convergence behavior of optimization algorithms}
We examined our algorithms for assessment of convergence
speed. We generated datasets artificially as follows. $K=50$
matrices $\vF_{k}\in\bR^{n\times n}$ are generated in which
each entry is drawn from the uniform distribution in the
interval $[-0.5,0.5]$. Then, we set
$\vA_{k}:=\vF_{k}\vF_{k}^\top$. The values of the variables
$y_{k}$ are randomly chosen from $\{\pm 1\}$ with same
probabilities. We set $\vb=\1$ and $\vc=\1/(\lambda K)$. We
exhaustively tested Cyclic, Perm, and Rand with the settings
of $\lambda=10^{-2},10^{-3},10^{-4}$ and $n=10,50,100$.

Figure~\ref{fig:logdet.primal}
demonstrates the convergence behavior of the
cyclic Dykstra algorithm and the two stochastic Dykstra
algorithm with various $\lambda$ and $n$. Here, one epoch is
called $K$ times projection onto a single half-space.
ISBD is employed as the objective function for learning the metric~$\vW$. 
The
objective gap is defined as the difference between the
current objective value and the minimum. In most of the
settings, the two stochastic Dykstra algorithms converged
faster than the cyclic algorithm. Especially when
$\lambda=10^{-4}$, the cyclic algorithm was too slow to use
it in practice.

\subsection{Generalization performance for pattern recognition}
We used the Brodatz texture dataset~\cite{randen1999filtering} containing 111
different texture images to examine the generalization
performance for texture classification. Each image has a size of
$640\times 640$ and gray-scaled. Images
were individually divided into four sub-images of equal size. One of
the four sub-images was picked randomly and used for
testing, and the rest of the images were used for training.

For each training image and each testing image, covariance
descriptors of randomly chosen $50$ were extracted from
$128\times 128$ patches. The covariance matrices are of
five-dimensional feature vectors
$\left[I,|I_{x}|,|I_{y}|,|I_{xx}|,|I_{yy}|\right]^\top$.
Then, $11,100(=111\times 2 \times 50)$ covariance
descriptors are obtained for training and testing,
respectively. For evaluation of generalized performance,
$k$-nearest neighbor classifier is used, where the number of
the nearest neighbors is set to three. We set $K=100\times
n_{c}$, $b_{\text{ub}}=0.05$, and $b_{\text{lb}}=0.95$.

We also examined the generalization performance for generic
visual categorization using the ETH-80 dataset~\cite{leibe2003analyzing} containing
$n_{c}=8$ classes. Each class has $10$ objects, each of which includes $41$
colored images. For every
object, $20$ images are randomly chosen and used for
training, and the rest of images are used for testing.

One covariance matrix is obtained from each image. Eight
features $\left[x, y, R, G, B,
|I_{x}|,|I_{y}|,|I_{xx}|,|I_{yy}|\right]^\top$ are obtained
from each pixel in an image.

\begin{figure*}[t!]
\begin{center}
  \begin{tabular}{ll}
    (a) Brodatz texture dataset
    & (b) ETH-80  dataset
    \\
    \includegraphics[width=0.4\textwidth]{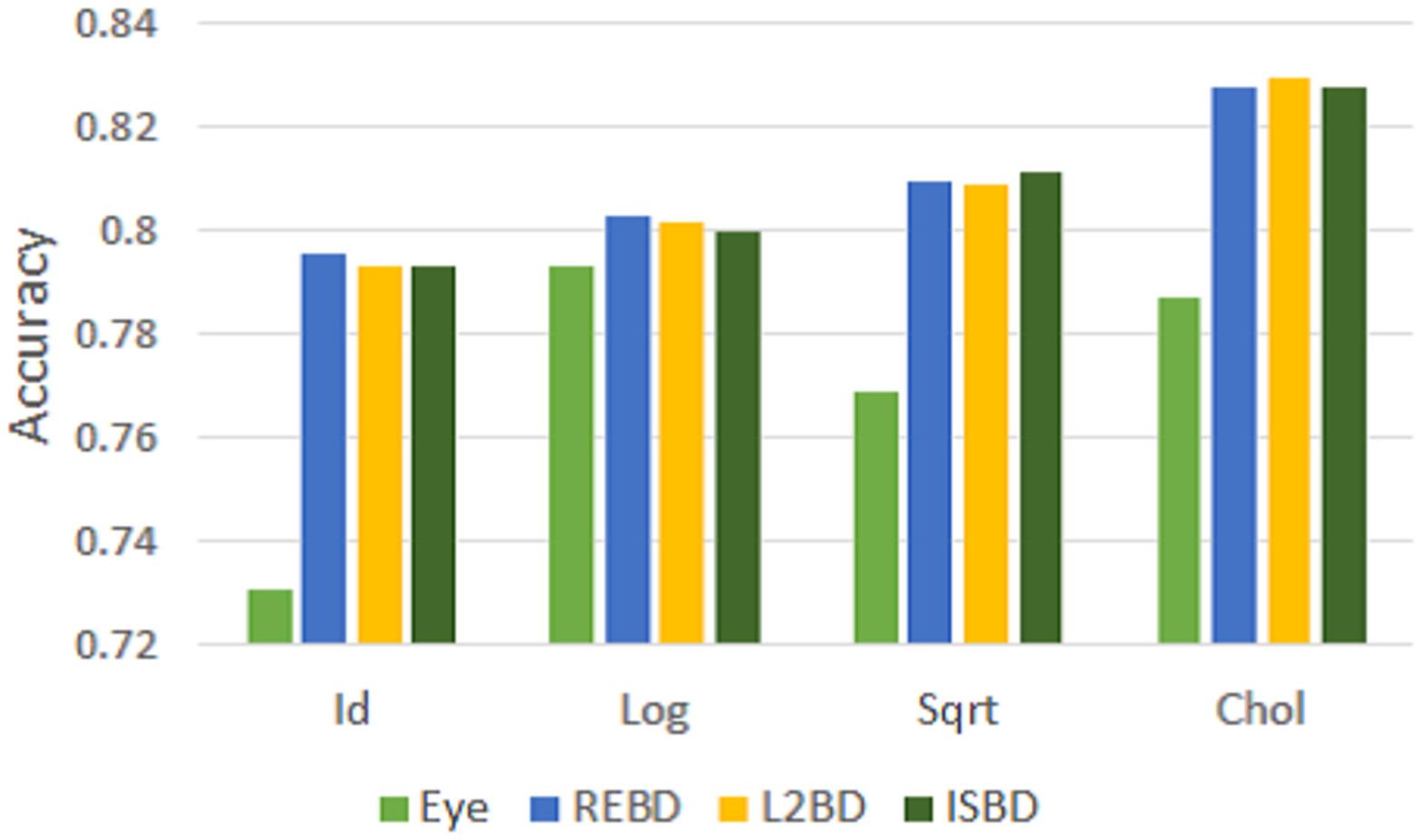}
    &
    \includegraphics[width=0.4\textwidth]{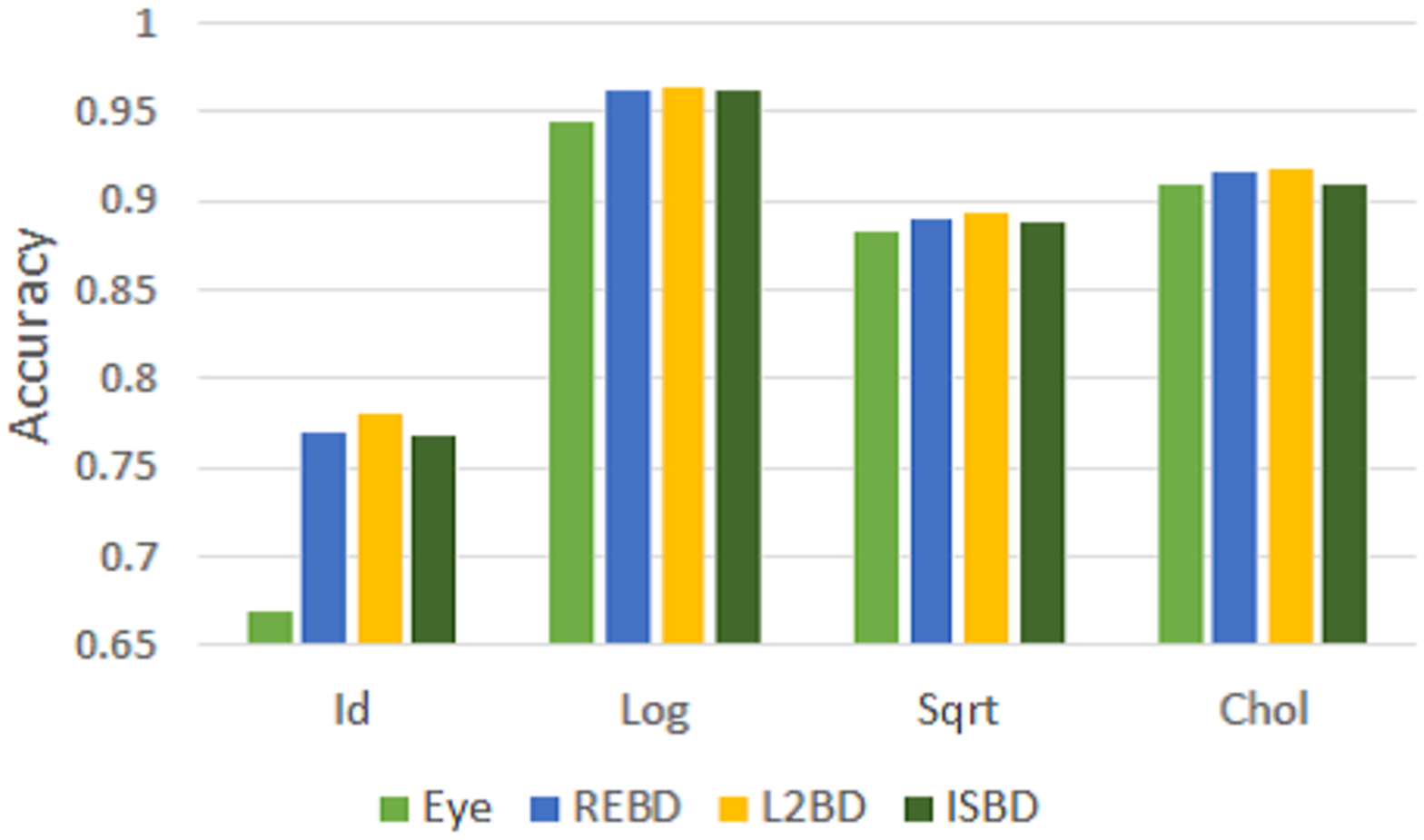}
\end{tabular}
\end{center}
\caption{Generalized performances for pattern recognition. \label{fig:accbars}}
\end{figure*}

We tried four types of $\vPhi$:
\textbf{Id}: $\vPhi(\X)=\X$,
\textbf{Log}: $\vPhi(\X)=\logm(\X)$,
\textbf{Sqrt}: $\vPhi(\X)=\X^{1/2}$,
\textbf{Chol}: $\vPhi(\X)=\text{chol}(\X)$. 
The parameter $\vW$ is determined
by the metric learning algorithms with
ISBD, L2BD, and REBD, to be compared with $\vW=\vI$ we denote as \textbf{Eye}. 
Note that $D_{\vPhi}(\cdot,\cdot;\vI)=D_{\vPhi}(\cdot,\cdot)$. 
Figure~\ref{fig:accbars} gives the accuracy bar plots for the two
multi-class classification problems. Whichever $\vPhi$ is
used, supervised metric learning improved the generalization
performances both for texture classification and for generic
visual categorization. For texture classification, the
Cholesky decomposition-based mapping $\text{chol}(\cdot)$
achieved the best accuracy, while the matrix logarithm-based mapping
$\text{logm}(\cdot)$ obtained the highest accuracy for generic image
categorization.

\section{Conclusions}
In this paper, we have devised several objective functions
for metric learning on positive semidefinite cone, all of
which can be minimized by the Dykstra algorithm. We have
introduced a new technique that performs each update
efficiently when the Dykstra algorithm is applied to the
metric learning problems. We have empirically demonstrated
that the stochastic versions of the Dykstra algorithm are
much faster than the original algorithm.

\section*{Acknowledgment}
This work was supported by JSPS KAKENHI Grant Number
26249075, 40401236. 
The last author would like to thank Dr. Zhiwu Huang for fruitful discussions.

\bibliographystyle{plain}

\end{document}